\documentclass{article}
\usepackage{spconf,amsmath,epsfig}
\usepackage{amssymb}
\usepackage{algorithm}
\usepackage{algpseudocode}
\usepackage{xcolor}

\let\OLDthebibliography\thebibliography
\renewcommand\thebibliography[1]{
  \OLDthebibliography{#1}
  \setlength{\parskip}{0pt}
  \setlength{\itemsep}{0pt plus 0.3ex}
}

\begin{document}\sloppy

\def\x{{\mathbf x}}
\def\L{{\cal L}}

\title{Domain-Generalized Textured Surface Anomaly Detection}
\name{
\parbox{\linewidth}{\centering
      Shang-Fu Chen$^{\ast \dag}$,
      Yu-Min Liu$^{\ast}$,
      Chia-Ching Lin$^{\ast}$,
      Trista Pei-Chun Chen$^{\dag}$,
      Yu-Chiang Frank Wang$^{\ast}$%
    }%
}

\address{$^{\ast}$ Graduate Institute of Communication Engineering, National Taiwan University, Taiwan\\
$^{\dag}$ Inventec Corporation, Taiwan}

\maketitle

%
\begin{abstract}
Anomaly detection aims to identify abnormal data that deviates from the normal ones, while typically requiring a sufficient amount of normal data to train the model for performing this task. Despite the success of recent anomaly detection methods, performing anomaly detection in an unseen domain remain a challenging task. In this paper, we address the task of domain-generalized textured surface anomaly detection. By observing normal and abnormal surface data across multiple source domains, our model is expected to be generalized to an unseen textured surface of interest, in which only a small number of normal data can be observed during testing. 
Although with only image-level labels observed in the training data, our patch-based meta-learning model exhibits promising generalization ability: not only can it generalize to unseen image domains, but it can also localize abnormal regions in the query image.
Our experiments verify that our model performs favorably against state-of-the-art anomaly detection and domain generalization approaches in various settings.
\end{abstract}

\section{Introduction}
Textured surface anomaly detection has been among a practical yet challenging task, which requires one to determine abnormal data from the normal ones. When it comes to real-world problems, e.g., quality control of industrial products, abnormal samples are generally difficult to be collected.
Therefore, existing solutions focus on training models which identify data that deviate from the learned distribution of normality as anomaly. With the recent advances of deep learning, a popular model choice is the autoencoder~\cite{zong2018deep, venkataramanan2020attention}, which trains to recover normal data samples and thus performs anomaly detection by the associated reconstruction loss.
To avoid the trained autoencoder to recover abnormal samples as well,~\cite{gong2019memorizing, tan2021trustmae} propose to learn memory banks to regularize the autoencoder, ensuring the data to be described by representative patterns. Despite the success of these reconstruction-based models, it is still a difficult task to perform anomaly detection in unseen data domains. Moreover, one cannot expect the derived distribution of normality to be applicable for different domains for anomaly detection.


Learning models from a single or multiple source domains, domain generalization~\cite{li2018domain, wang2020eis, li2018domain, yue2019domain} aims to leverage this model to unseen target domains for solving the same learning task. A straightforward yet naive baseline approach is to aggregate training samples from all source domains to learn a single model.
To further improve the generalization capability, \cite{li2019episodic} designs an episodic learning procedure that simulates the domain shift observed during training for deriving a domain generalized model. \cite{wang2020eis} argues that a properly learned domain generalization model would discover the image intrinsic properties, which are irrelevant to the data domains. Thus, self-supervised auxiliary learning tasks are introduced to prompt the learning of their models.

\begin{figure}
\begin{center}
\includegraphics[width=\linewidth]{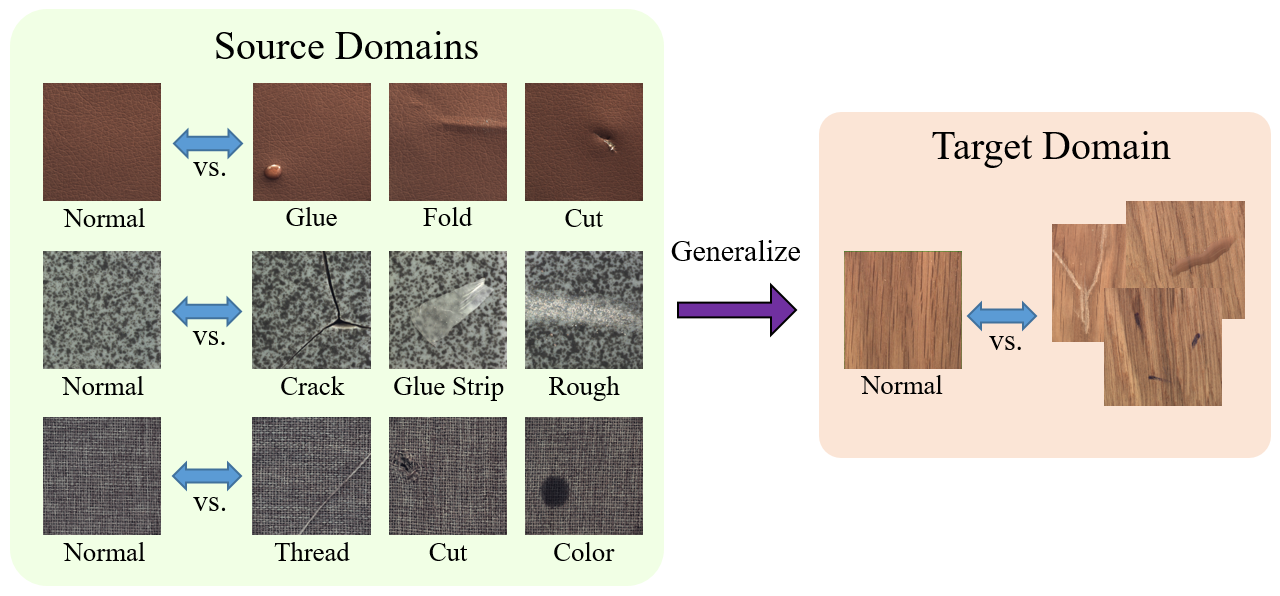}
\end{center}
\vspace{-0.4cm}
\caption{Illustration of domain-generalized anomaly detection. By observing normal and abnormal data multiple source domains, the learned model needs to generalize to perform anomaly detection in an unseen target domain where only a small amount of normal images are available during testing.}
\label{teaser}
\end{figure}

Although the recent success of domain generalization has benefited a wide range of computer vision applications, it would not be feasible for anomaly detection if no normal data is presented in the domain of interest for a standard reference of normality. Thus, if one expects to address such tasks in an unseen target domain, at least a number of normal data in that domain needs to be observed during testing. In other words, existing domain generalization methods like~\cite{li2019episodic, wang2020eis, li2018domain, yue2019domain} cannot be easily applied for solving the above problem.

To address the above concerns and challenges, we tackle the task of domain-generalized textured surface anomaly detection in this paper. That is, with collection of training normal and abnormal data from existing source domains, i.e., textured surface, we aim to learn a model which can be generalized to detect abnormal data in \textit{unseen} target domain of interest. The problem definition and the idea of our work can be seen in Figure~\ref{teaser}. It is worth noting that, during the inference stage, only a small amount of normal samples are available for the target domain of interest, which follows the settings of most anomaly detection approaches~\cite{ zong2018deep, gong2019memorizing, tan2021trustmae, akcay2018ganomaly, schlegl2019f}. However, without the requirement of model fine-tuning, the trained model can be directly applied to such data domains which are not seen during training.

To highlight the technical novelty of our work, we introduce a meta-comparer module that learns to compare textured surface data for anomaly detection across multiple source domains. We take the normal image data as the reference images, and perform patch-level co-attention on the query-reference image pairs during training. With only image-level labels observed (i.e., normal and abnormal data), the above co-attention mechanism guides the meta-comparer to identify the normality of the query input, resulting in both image-level and patch-level anomaly detection. Since our model is trained to compare image pairs across different source domains in a meta-learning fashion, the learned model is shown to exhibit promising generalization ability for unseen data domains.

Our contributions can be summarized as follows:
\begin{itemize}
    \item We address the task of domain-generalized textured surface anomaly detection. Given a number of normal (reference) images in unseen target domains, our model is able to perform anomaly detection accordingly.
    \item We propose a meta-learning framework that learns to compare images in a query-reference pair across multiple source domains. Therefore, our learned model is able to generalize to unseen image domains for identifying abnormal images.
    \item With only image-level labels observed, a co-attention mechanism across query-reference image pairs is introduced, which guides our meta-comparer to realize not only image-level anomaly detection but also patch-level anomaly localization.
\end{itemize}

\section{Method}
\begin{figure*}
\begin{center}
\includegraphics[width=0.8\linewidth]{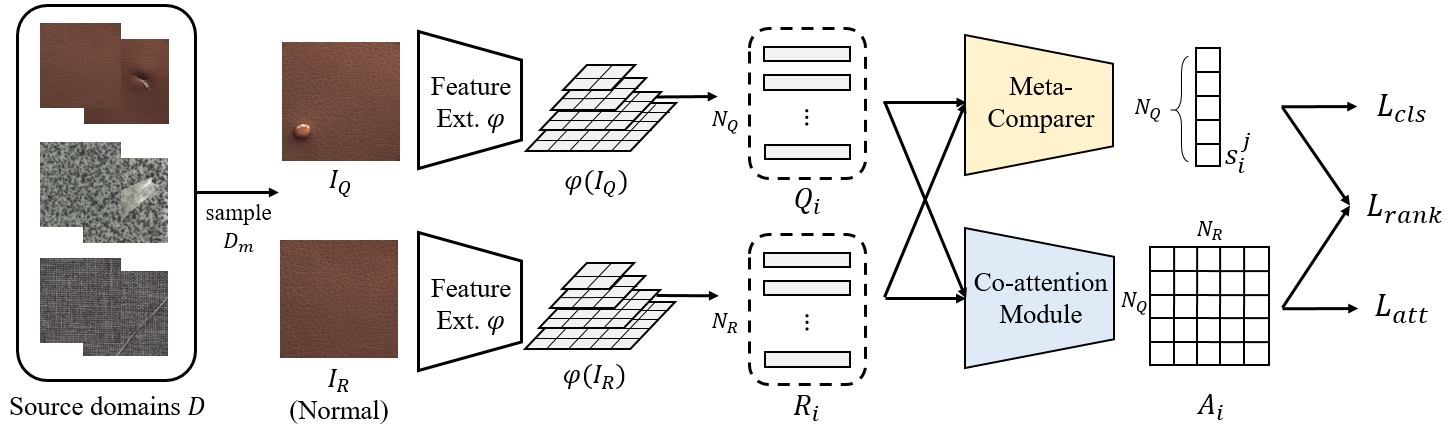}
\end{center}
\vspace{-0.3cm}
\caption{Overview of our proposed model, which consists of a feature extractor $\phi$, a co-attention module, and a meta-comparer. The feature extractor aims to derive multi-scale patch-based features for both query and reference images. The co-attention module observes patches from such query-reference image pairs, guiding the meta-comparer to perform anomaly detection and localization.}
\label{archi}
\end{figure*}
For the sake of clarification, we first define the notations and setting considered in this paper. We observe image data from $M$ source domains $D = [D_1, ..., D_M]$ at the training stage. Each $D_m$ contains image-label pairs $(x^n_m, y^n_m)$, in which $y^n_m$ is either 0 or 1 representing normal or abnormal labels. Note that we assume that only image-level labels are available during training, i.e., no pixel-level anomaly ground truth can be observed. Our goal is to train a model using $D$ in a meta-learning manner, and have this model generalized to perform anomaly detection on an unseen target domain $D_{M+1}$ where only a number of normal images are available during testing. 

The overview of our proposed framework is depicted in Figure~\ref{archi}. From this figure, we see that our learning model contains three components: a feature extractor, a co-attention module, and a meta-comparer. The feature extractor $\phi$ aims to derive multi-scale features from query and reference (i.e., normal) images. The co-attention module observes query-reference image pairs, resulting in proper patch-level supervision, which guides the meta-comparer for producing the resulting anomaly score. By sampling different source domains $D_m$ during the training stage, our meta-comparer learns to compare query-reference image data in a meta-learning fashion. In the following sections, we will detail the functionality and design of each module.

\subsection{Multi-Scale Feature Extraction}
In our proposed framework, the feature extractor is expected to extract the features from the query image $I_Q$ and the reference image $I_R$ from a domain of interest. We note that, while the query images are with labels $y=1$ or $0$ during training, we only consider the normal one as the reference for both training (from multiple source domains) and testing (on unseen target domains). 
Following techniques utilized for object detection (e.g.,~\cite{lin2017feature},~\cite{tan2020efficientdet}), we consider multi-scale features from image data for aiming at not only to recognize the abnormal query input, but also for the purpose of identifying the defect regions. More precisely, we apply the bi-directional feature pyramid network (BiFPN) proposed by~\cite{tan2020efficientdet} to produce a feature pyramid with multiple resolutions.

Take the query image $I_Q$ as an example, the feature extractor $\phi$ extracts a feature pyramid containing $L$ feature maps with different resolutions/scales. The associated multi-scale features are denoted as $\phi_1(I_Q), \phi_2(I_Q), ..., \phi_L(I_Q)$, where $\phi_i(I_Q)$ represents the query feature at scale level~$i$. Let $N_Q$ denotes the number of patches sampled from $\phi_i(I_Q)$, we thus have a set of patch-based representations $Q_i = \{q^1_i, q^2_i, ..., q^{N_Q}_i\}$ for the query image $I_Q$ at scale level $i$. Similarly, we have $R_i = \{r^1_i, r^2_i, ..., r^{N_R}_i\}$ as the set of patch-based representations for the reference image $I_R$ at scale level $i$, where $N_R$ denotes the number of sampled patches. For the detailed process of the multi-scale feature extraction, please refer to the supplementary materials.
\subsection{Image-Level Anomaly Detection}
With patch features extracted from the query and reference images, we now explain how we train our feature extractor and meta-comparer for performing image-level anomaly detection.
For the $j$-th query patch $q^j_i$ at scale level $i$, the meta comparer is utilized to calculate its largest query-reference anomaly score $s^j_i$ as:
\begin{equation}
\begin{aligned}
\label{eq:s_i}
s^j_i = \max\limits_{k = [1, ..., N_R]}\text{MLP}([q^j_i, r^k_i]),
\end{aligned}
\end{equation}
where MLP denotes a multilayer perceptron module with Sigmoid activation functions deployed. It can be expected that, if the query image $I_Q$ is abnormal, at least one query patch $q^j_i$ would remarkably deviate from the reference patches, and thus the value of the corresponding $s^j_i$ would be close to 1.



With the above observation, we define the image-level classification loss (under supervision of $y$) as follows:
\small
\begin{equation}
\begin{aligned}
\label{eq:l_cls}
L_{cls} = -\sum^L_{i = 1}y\log(\max\limits_{j}(s^j_i))+(1-y)\log(1-\max\limits_{j}(s^j_i)).
\end{aligned}
\end{equation}
\normalsize
In the above equation, $\max\limits_{j}(s^j_i)$ calculates and outputs the largest anomaly score from the query patches at scale $i$, which sums over all $L$ scales for the resulting loss output.
\subsection{Patch-level Anomaly Localization}
In addition to image-level anomaly detection, the introduced co-attention module in our framework of Figure~\ref{archi} allows us to perform the same task at patch level. Therefore, localization of abnormal surface regions can be achieved via patch-level anomaly detection with only image-level label $y$ required.\\

\noindent \textbf{Co-attention on query-reference image pairs} The co-attention module first maps the query-reference patch pairs (i.e., $q^j_i$ and $r^k_i$) at scale $i$ into a shared latent space, followed by the calculation of cosine similarity between them. This produces a co-attention matrix $A_i \in \mathbb{R}^{N_Q \times N_R}$, which can viewed as an affinity matrix of $Q_i$ and $R_i$ at scale $i$, reflecting the similarity between the associated patch pairs. 

Similar to image-level anomaly detection, we observe that if the query image $I_Q$ is abnormal, then there would exist at least one query patch $q^j_i$ which would be distinct from the reference ones $r^k_i$. That is, if $y=1$ for the query, we expect at least one query-reference patch pair in $A_i$ resulting in a low similarity score. On the other hand, if $y=0$ for the query, every query-reference pair is expected to produce a large similarity score. Thus, by normalizing the attention matrix $A_i$ to $[0, 1]$, we introduce and calculate the following attention loss $L_{att}$ across image scales,
\small
\begin{equation}
\begin{aligned}
\label{eq:l_att}
L_{att} = -\sum^L_{i = 1}y\log(1-\min\limits_{j, k}(a^{j, k}_i))+(1-y)\log(\min\limits_{j, k}(a^{j, k}_i)),
\end{aligned}
\end{equation}
\normalsize
where $\min\limits_{j, k}(a^{j, k}_i)$ denotes the query-reference patch pair at scale $i$ with the minimum similarity score. Note that $j$ and $k$ are the patch indices for the query and reference images, respectively.


With the above co-attention mechanism, we calculate the co-attention score for $q^j_i$ as $a^j_i = \max\limits_{k}(a^{j, k}_i)$. In the formula, $a^j_i$ calculates the score between $q^j_i$ and every reference patch, and outputs the score with the most similar reference patch as the attention guidance. It can be expected that, if the query patch $q^j_i$ is abnormal, such $a^j_i$ scores would be close to 0 (and vice versa). Therefore, the co-attention score $a^j_i$ can be a patch-level guidance for the query patch $q^j_i$.\\

\noindent \textbf{From patch-level co-attention to anomaly localization}
In our proposed framework, patch-level anomaly detection is achieved by sampling pairs of patches $q^u_i, q^v_i$ from a query $Q_i$ at scale $i$, followed by the meta-comparer to produce their patch-level anomaly scores $s^u_i, s^v_i$ under the supervision of $y$ and the guidance of the aforementioned co-attention outputs. Inspired by~\cite{wang2020learning}, we introduce a patch-level anomaly ranking loss $L_{rank}$ for the sampled query patch pairs as follows,
\begin{equation}
\begin{aligned}
\label{eq:l_rank}
&L_{rank} = \sum\limits_{i=1}^L\sum_{q^u_i, q^v_i \in Q_i}w^{uv}_i\max(0, 1-\sigma(s^u_i-s^v_i)),\\
&\text{where\hspace{0.3cm}} w^{uv}_i = \lambda(\exp(|a^u_i - a^v_i|)-1),\\
&\text{and\hspace{0.3cm}} \sigma = -\textit{sgn}(a^u_i - a^v_i).
\end{aligned}
\end{equation}
Note that $\lambda$ is a scaling factor, and $sgn$ indicates the sign function that extracts the sign of a real number.
From equation~\eqref{eq:l_rank}, we see if both $q^u_i, q^v_i$ are the normal patches, both co-attention scores $a^u_i$ and $a^v_i$ would be large, and the corresponding $w^{uv}_i$ is close to 0. This would result in the ranking loss $L_{rank}$ close to 0 as well. Similarly, if both are the abnormal ones, we have similar yet small $a^u_i$ and $a^v_i$ values, which produces small $w^{uv}_i$ regularizing the ranking loss as well. Finally, and most importantly, if only one of $q^u_i$ and $q^v_i$ is abnormal, we would observe very different co-attention score $a$ and thus produce a large $w^{uv}_i$. If the co-attention score $a^u_i$ is less than $a^v_i$, the corresponding anomaly score $s^u_i$ should be larger than $s^v_i$. To ensure this property, the variable $\sigma$ verifies the order of $s^u_i$ and $s^v_i$ according to their corresponding co-attention score $a$. With the goal of anomaly localization, the above objective allows us to automatically identify the query patch which deviates not only from the reference ones but also from the remaining ones in the query.




\begin{figure}
\begin{center}
\includegraphics[width=0.9\linewidth]{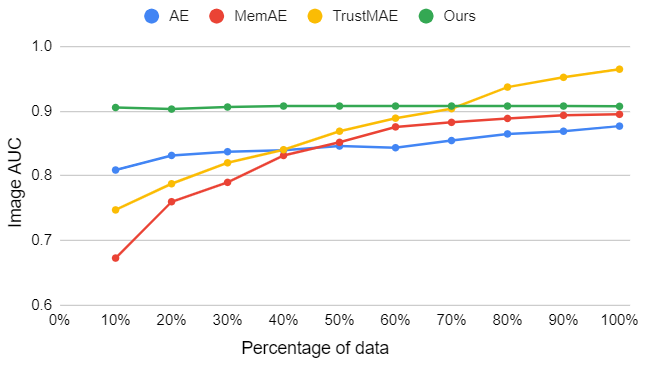}
\end{center}
\vspace{-0.35cm}
\caption{Average image-level AUC for anomaly detection over the 5 textures of MVTec-AD, with different percentage of normal reference images from the target domain.}
\label{ad_dec}
\end{figure}
\subsection{Domain-Generalized Anomaly Detection}

With the introduced image-level detection and patch-level localization discussed above, we now explain how our proposed framework is trained to exhibit additional domain generalization ability. During training, by sampling query-reference image pairs $(I_Q, I_R)$ from multiple source domains, we enforce the meta-comparer and the co-attention module for learning to compare image data by applying equation~\eqref{eq:l_cls} and equation~\eqref{eq:l_att}, disregard of the data domain distributions. Moreover, by sampling different query patch pairs $q^u_i$ and $q^v_i$ in Equation~\eqref{eq:l_rank}, our meta-comparer further performs the above learn-to-compare scheme in the patch level. Therefore, our model is expected to learn a generalized capability of comparing image data. The full objectives of our model and the detailed training process are summarized in the Algorithm A of our supplementary materials.

As for the inference stage, we apply our model to an unseen target domain $D_{M+1}$ with a small amount of normal samples are presented. 
We first calculate the patch-level anomaly score $s^j_i$ for each extracted query patch $q^j_i$. If there exists a patch with defect regions at any scale, the query image is considered to be abnormal. Therefore, the image-level prediction $\hat{y}(I_Q)$ for $I_Q$ can be calculated by simply taking the maximum anomaly scores among all query patches $q^j_i$:
\begin{equation}
\label{eq:im_pred}
\hat{y}(I_Q) = \max(\{s^j_i\}) \quad \forall i, j.
\end{equation}
\noindent If localization of defect regions would be needed, we can calculate the anomaly score for each pixel $p$ in $I_Q$ according to patch-level anomaly scores across multiple scale levels. This is realized by taking the maximum anomaly scores among all query patches containing pixel $p$:
\begin{equation}
\label{eq:patch_pred}
\hat{y}(p) = \max(\{s^j_i\}) \quad \forall i, j \text{ such that } p \in q^j_i.
\end{equation}


\begin{figure}
\begin{center}
\includegraphics[width=0.9\linewidth]{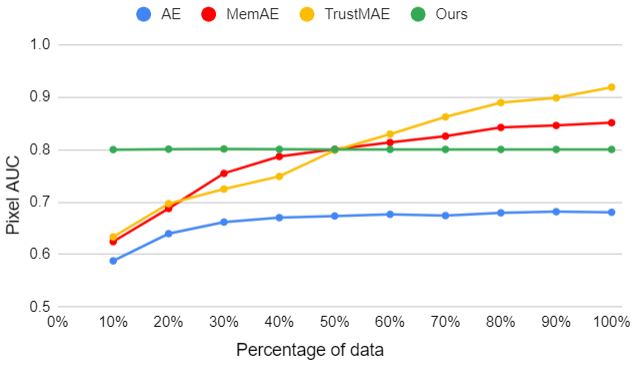}
\end{center}
\vspace{-0.35cm}
\caption{Average pixel-level AUC for anomaly localization over the 5 textures of MVTec-AD, with different percentage of normal reference images from the target domain.}
\label{ad_loc}
\end{figure}
\section{Experiments}
\begin{figure}
\begin{center}
\includegraphics[width=0.9\linewidth]{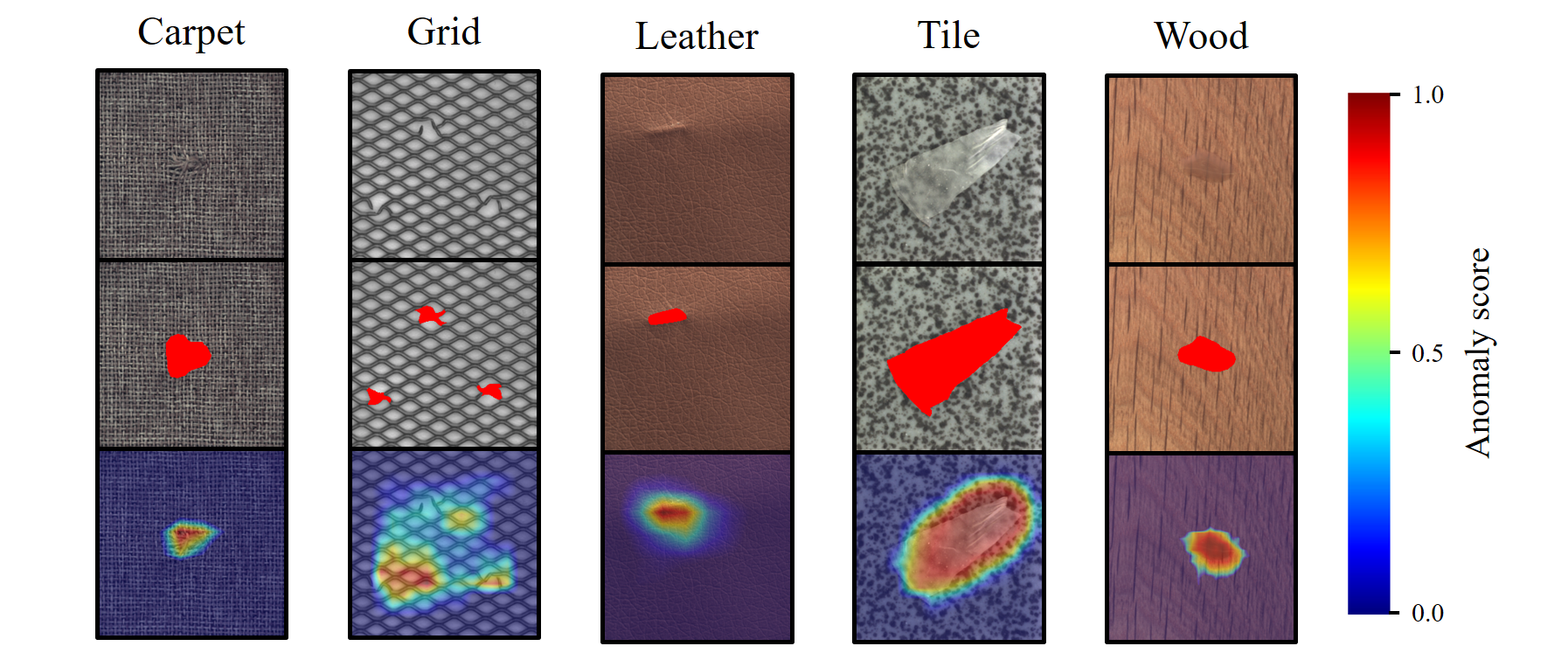}
\end{center}
\vspace{-0.6cm}
\caption{
Visualization of anomaly localization of our method on MVTec-AD. The top row shows input abnormal images, the middle row indicates the ground truth defect regions, and the bottom row shows our anomaly localization results.}
\vspace{-0.2cm}
\label{visual}
\end{figure}

We evaluate our proposed framework on MVTec-AD~\cite{bergmann2019mvtec} and BTAD~\cite{mishra2021vt} datasets. The MVTec-AD dataset consists of 3,629/1,725 training/testing images from 5 texture and 10 object products. In this paper, we consider the texture products of MVTec-AD for textured surface anomaly detection, i.e., Carpet, Grid, Leather, Tile and Wood, as showed in Figure~\ref{visual}.
For these 5 texture types, we follow recent domain generalization approaches~\cite{li2019episodic} and~\cite{wang2020eis} and do leave-one-out evaluation, in which only one texture is selected at a time as the target domain at the inference stage, while the remaining four textures are used as the source domains during training. Following previous works~\cite{gong2019memorizing, tan2021trustmae, mishra2021vt, bergmann2020uninformed}, we evaluate the models using the area under the receiver operating characteristic curve (AUC). 

As for the BTAD dataset, it consists of 2,250/291 normal/abnormal images from 3 industrial products. The image data from this dataset would serve as the (unseen) target domains for testing in experiments for the cross-dataset settings, which would further verify the effectiveness of our propose method for domain-generalized anomaly detection. The implementation details and the results of the cross-dataset experiments are demonstrated in the supplementary materials.

\subsection{Quantitative Results}
In our experiments, we compare our model with a number of recent anomaly detection (AD) and domain generalization (DG) approaches. For fair comparisons, we adopt the same pre-trained ResNet-18 feature extractor for all the methods considered. Moreover, to comply with our domain-generalized anomaly detection setting, we allow AD and DG methods to take normal image data from the target domain as additional inputs during the inference stage as well.\\
\begin{table}[t]
\footnotesize
\centering

\begin{tabular}{ c | c  c  c  c  c | c }
	\hline
	            &   Carpet  &   Grid    &   Leather &   Tile    &   Wood    &   Avg. \\
    \hline
    AGG~\cite{li2019episodic}    &   0.875   &   0.628   &   0.981   &   0.886   &   0.852   &   0.845   \\
    Epi-FRC~\cite{li2019episodic} &   0.916   &   0.640   &   0.995   &   0.947   &   0.909   &   0.881   \\
	EISNet~\cite{wang2020eis} &   \textbf{0.991}   &   0.662   &   1.000   &   0.850   &   \textbf{0.986}   &   0.898   \\
    \hline
	AGG+    &   0.891   &   0.608   &   0.992   &   0.912   &   0.865   &   0.854   \\
    Epi-FRC+ &   0.916   &   0.725   &   1.000   &  0.951   &   0.941   &   0.907   \\
	EISNet+ &   0.982   &   0.728   &   1.000   &   0.858   &   0.979   &   0.909   \\
	\hline
    Ours        &   0.943   &   \textbf{0.730}   &   1.000  &   \textbf{0.956}   &   0.962   &   \textbf{0.918}   \\
	\hline
\end{tabular}

\caption{Domain-generalized anomaly detection on MVTev-AD with the leave-one-domain-out setting in terms of the average image-level AUC. Note that the $+$ notation denotes the modified versions for existing DG approaches (i.e., with learn-to-compare scheme introduced).}
\label{ad_dg}
\end{table}

\noindent \textbf{Comparisons to existing AD Approaches}
We compare our model AD approaches, including an autoencoder baseline~\cite{gong2019memorizing} as well as two state-of-the-art methods of MemAE~\cite{gong2019memorizing} and TrustMAE~\cite{tan2021trustmae}. 
We follow the officially-released code and the instruction presented in the paper to implement the above methods.
A common limitation of existing AD approaches is that a sufficient amount of training data from the domain of interest would be needed. As noted in previous sections, existing anomaly detection approaches use all the available normal images from the target domain \textit{for training}.
On the other hand, our model does not require any normal image data in the target domain for training, and only observes such data as references \textit{during inference}. We compare our method to these AD approaches on MVTec-AD with same amount of target normal samples are observed. With the aforementioned leave-one-domain-out setting, we control the percentage of the amount of target normal samples and compare the average image-level AUC for anomaly detection and pixel-level AUC for anomaly localization in Figure~\ref{ad_dec} and Figure~\ref{ad_loc}, respectively.
As can be seen from these two figures, existing AD approaches required a sufficient amount of normal training data in the target domain (e.g., above 60 or 70\% of the target-domain normal data available) to achieve satisfactory performances, while our method consistently outperformed such methods especially even with only 10\% (i.e., about 25 images) of such data were observed.
This is expected since our proposed model only utilizes the target normal samples as reference during inference. Therefore, the performance of our method is not sensitive to the amount of such data, which would be preferable for practical uses.



\noindent \textbf{Compare with existing DG Approaches}
As for recent DG approaches, we consider a baseline of simple aggregation of AGG~\cite{li2019episodic}, and two state-of-the-art methods of Epi-FCR~\cite{li2019episodic} and EISNet~\cite{wang2020eis} for comparisons. We note that, existing DG models generally make prediction solely based on the query image, not in the learn-to-compare fashion as ours does. Thus, for fair comparison, we additionally modify the above DG approaches to take query-reference pairs as training inputs, and such modified versions are denoted as $+$ in our results presented in Table~\ref{ad_dg}. We also note that, for fair comparisons, all target-domain normal reference images are utilized for all DG methods and ours in the experiments.

From the results listed in Table~\ref{ad_dg}, we see that our method performed favorably against existing DG approaches (for both the original and the modified learn-to-compare versions) over all 5 texture categories in terms of the average AUC. It is interesting to point out that, from the results shown in this table, the modified versions of recent DG approaches (i.e., with learn-to-compare mechanism introduced) were shown to produce improved performances when comparing to their original versions. This suggests that by a properly designed learn-to-compare scheme as ours is, the anomaly detection model can be expected to generalize to unseen target domains. 
It can be seen that our model outperforms all existing DG approaches by a large margin. It is expected since our model explores the relationships between patch features for detecting sophisticated defects, while the above methods only consider image-level features for anomaly detection.
\subsection{Visualization of Anomaly Detection}
As discussed in Section 2, our proposed model not only performs anomaly detection but also exhibits abilities in identifying abnormal regions with only \textit{image-level} labels observed during training. We show the visualization results for anomaly localization in Figure~\ref{visual}. The top row of this figure shows input images containing defects; the middle row are the ground truth regions of defects (annotated in red); the bottom row shows the anomaly localization results predicted by our model. It can be seen that, from the example results shown in this figure, our model is able to accurately localize either small defects (in Carpet and Wood) or large defects (in Tile). It is also worth noting that, existing AD or DG approaches cannot easily address such anomaly localization without proper pixel-level guidance.
\subsection{Further Analysis and Remarks}
\begin{table}
\footnotesize
\centering

\begin{tabular}{ c | c  c  c  c  c }
	\hline
	            &   Carpet  &   Grid    &   Leather &   Tile    &   Wood  \\
    \hline
	Carpet      &   0       &   4.424   &   1.34    &   1.763   &   1.526 \\
    Grid        &   4.424   &   0       &   3.966   &   4.646   &   4.409 \\
	Leather     &   1.34    &   3.966   &   0       &   1.916   &   1.601 \\
	Tile        &   1.763   &   4.646   &   1.916   &   0       &   2.032 \\
	Wood        &   1.526   &   4.409   &   1.601   &   2.032   &   0     \\
	\hline
    Average     &   1.811   &   3.489   &   1.765   &   2.071   &   1.914 \\
	\hline
\end{tabular}

\caption{FID scores between each data domain pairs in MVTec-AD, which imply the difficulty expected for domain-generalized anomaly detection.}
\vspace{-0.2cm}
\label{fid}
\end{table}
To further verify the capability and point out the limitation of our domain generalization method, we quantitatively assess the domain differences between different texture categories from MVTec-AD, reflecting the expected DG difficulty for the associated target domain. To analyze the above issue, we apply the Fréchet Inception Distance (FID) score introduced by~\cite{heusel2017gans} to calculate the differences between each texture/domain pairs and list the results in Table~\ref{fid}.

From Table~\ref{fid}, we see that the Grid texture generally has larger FID scores (average 3.489) than those of other texture types, suggesting that the distribution of Grid deviates more drastically from those of other texture category data. This observation is consistent to the AUC results shown in Table~\ref{ad_dg}, where all DG methods (including ours) did not report comparable performances when Grid was the unseen target domain of interest. On the other hand, since the average FID of Leather is the smallest, the knowledge learned by the model from other source domains is expected to generalize data in this domain, which also explains why all DG methods reported highest AUC performances in Table~\ref{ad_dg}. In other words, while we claim that our model can be generalized to unseen target domain for anomaly detection, the performance drop would be expected if the target domain data distribution would be very different from those of source domain data.

\section{Conclusion}
In this paper, we tackle the task of domain-generalized anomaly detection. With only image-level labels observed for multiple source domains, our model learns to compare images in query-reference pairs across the above data domains during training. With the co-attention mechanism introduced, our model learns to compare and identify abnormal image data and the associated defect regions, and it is shown to achieve promising performances on anomaly detection and localization for unseen target domain data.\\

\noindent\textbf{Acknowledgement}
We thank National Center for High-performance Computing (NCHC) and Inventec Cooperation for providing computational and storage resources.
%
%

\bibliographystyle{IEEEbib}
\bibliography{egbib}

\end{document}